\documentclass{article}

\usepackage{microtype}
\usepackage{graphicx}
\usepackage{subfigure}
\usepackage{booktabs} 

\usepackage{hyperref}

\usepackage[accepted]{icml2020}

\usepackage{amsmath}
\usepackage{amsthm}
\newtheorem{theorem}{Theorem}[section]

\icmltitlerunning{Recourse for Humans}

\begin{document}

\twocolumn[

\icmltitle{Learning Recourse Costs from Pairwise Feature Comparisons}



\icmlsetsymbol{equal}{*}

\begin{icmlauthorlist}
\icmlauthor{Kaivalya Rawal}{goo}
\icmlauthor{Himabindu Lakkaraju}{ed}
\end{icmlauthorlist}

\icmlaffiliation{goo}{Institute for Applied Computational Science, Harvard University, Cambridge, MA, USA}
\icmlaffiliation{ed}{Harvard Business School, Harvard University, Boston, MA, USA}

\icmlcorrespondingauthor{Kaivalya Rawal}{kaivalyarawal45@gmail.com}

\icmlkeywords{Machine Learning, ICML, Recourse, Counterfactuals, Preference Learning, Bradley-Terry models}

\vskip 0.3in
]



\printAffiliationsAndNotice{}  

\begin{abstract}
This paper presents a novel technique for incorporating user input when learning and inferring user preferences. When trying to provide users of black-box machine learning models with actionable recourse, we often wish to incorporate their personal preferences about the ease of modifying each individual feature. These recourse finding algorithms usually require an exhaustive set of tuples associating each feature to its cost of modification. Since it is hard to obtain such costs by directly surveying humans, in this paper, we propose the use of the Bradley-Terry model to automatically infer feature-wise costs using non-exhaustive human comparison surveys. We propose that users only provide inputs comparing entire recourses, with all candidate feature modifications, determining which recourses are easier to implement relative to others, without explicit quantification of their costs. We demonstrate the efficient learning of individual feature costs using MAP estimates, and show that these non-exhaustive human surveys, which do not necessarily contain data for each feature pair comparison, are sufficient to learn an exhaustive set of feature costs, where each feature is associated with a modification cost.
\end{abstract}

\vspace{-0.3in}
\section{Introduction}
\vspace{-0.05in}
As machine learning models have come to be deployed in more and more fields concerning daily human life, interest has grown commensurately in understanding these models and making them interpretable to end users. One way to do this is to construct a counterfactual explanation \cite{wachter2018a} which represents a small modification to the model inputs in order to obtain a desirable output. When they incorporate user preferences regarding the mutability of features, these explanations can be thought of as providing actionable recourse, as they enable the users of the models learn about the modifications required to obtain desirable outcomes \cite{Ustun_2019}.

In high stakes decision settings such as credit scoring, processing bail applications, or making hiring decisions, applicants often seek recourse to correct unfavourable predicted outcomes for the future. In these scenarios, since there can be multiple possible recourses for each individual, feasibility considerations, user preferences, and heuristics to minimize the size of the proposed modifications are used to guide the search for appropriate recourses \cite{FACE,Pawelczyk_2020,joshi2019realistic}. Recourse search algorithms thus return the best possible recourse based on these considerations by performing a search over the feature-space of the model.

\begin{figure}[ht]
\begin{center}
\centerline{\includegraphics[width=\columnwidth]{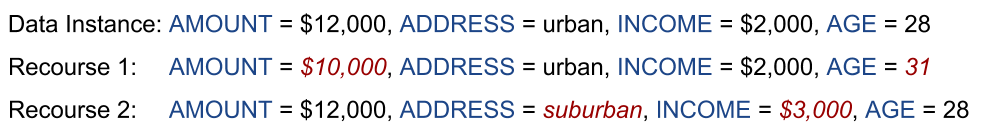}}
\caption{An example of a data instance that is denied a loan by a black box model, and two potential recourses that, if implemented, would get the loan application approved. Numeric costs on the ease of modification for each of the four features for Recourse 1 and Recourse 2 are essential to be able to determine which recourse is better for the user overall.}
\label{fig:rec-eg}
\end{center}
\vskip -0.2in
\end{figure}

 Specific techniques used by various recourse search algorithms are beyond the scope of this paper, however a recourse search algorithms can be generalized to involve users directly deciding the relative costs of modifying each feature. For example, in figure 1 above, if we do not use any proxy reliant on data distributions or heuristics, human input is necessary to gauge whether Recourse 1 is easier than Recourse 2. However, it is hard for users to place exact numbers on how easy it is to modify one feature versus another. For instance, it can be notoriously difficult to survey people for the value of $n$ from the following question: \textit{Modifying address is 'n' times easier than modifying income. What is an appropriate value of 'n'?}. By comparison, the question: \textit{Is modifying address easier than modifying income?} is much more human friendly, as people are much more comfortable ranking, ordering, and comparing quantities than explicitly assigning numeric values to them \cite{chaganty2016much}.

\textbf{Contributions:}
In this paper, we first show that comparisons are insufficient to disambiguate potential recourses, and it is essential to have numeric costs associated with each feature. We then demonstrate that it is possible to learn such numeric feature costs for use in recourse search algorithms without directly surveying humans. We propose the use of the well known Bradley-Terry model to convert the pairwise feature comparisons provided by users into numeric feature modification costs for use in recourse generation algorithms. We show that by exhaustively surveying users with pairwise comparisons over every possible set of features, we can exactly infer the numeric feature modification costs implicit in their minds. Finally, in order to improve the real world applicability of our system, we demonstrate via simulations that the pairwise comparisons surveyed from users do not need to be exhaustive (over every pair of features). By only collecting data comparing entire recourses with each other, with each recourse comprising of a set of distinct individual feature modifications,, we are able to retrieve the "ease-of-modification" costs of each individual feature.


\vspace{-0.1in}
\section{Methods}
\vspace{-0.05in}
In this section we provide a brief description of our solution leveraging the Bradley-Terry model, which finds the MAP estimates of the ease-of-modification costs for every feature used in the model, using pairwise comparison data surveyed from human users. We initially assume this survey to be exhaustive and contain a human response indicating the easier-to-modify feature between all possible pairs of features. Later we relax this condition and instead only use human inputs to decide which of two recourses as a whole (potentially consisting of multiple feature modifications each) is easier to implement.

\textbf{The Bradley-Terry Model:} Consider a trained black-box model $\mathcal{B}$, which operates upon vectors consisting of features $\mathcal{F}$ to output a binary classification $+1$ or $-1$. Any given recourse modifies some features $R \subseteq \mathcal{F}$ to change the classifier prediction. The Bradley-Terry model is a probabilistic model based upon Luce's choice axiom \cite{BT1,BT2}. It works by assigning a strength parameter $\beta_f$ to each feature $f \in \mathcal{F}$ employed by the model. Let $f>g$ denote that feature $f$ is easier to modify than feature $g$ for any two arbitrary features $f, g \in \mathcal{F}$, where $f \neq g$. The probability $p_{f>g}$ that a feature $f$ is easier to modify than $g$ is then defined using the strength parameters $\beta_f$ and $\beta_g$ of the two features, respectively, as: 
\begin{equation}
    p_{f>g} = \frac{e^{\beta_f}}{e^{\beta_f} + e^{\beta_g}}
\end{equation}
The Bradley-Terry model thus probabilistically captures comparison relationships between features, and maintains transitive relations (that is, if feature $f$ is probably easier to modify than feature $g$, which is probably easier to modify than feature $h$, then feature $f$ is also probably easier to modify than feature $h$). We can survey users with different pairs of features $f, g \in \mathcal{F}$ asking which of the two is easier to modify. In our probabilistic setting, we presume that the result of surveying the comparison between the same two features $f$ and $g$ can provide differing results $f>g$ and $g>f$ at different times. If this survey is exhaustive and consists of all possible pairs of features, then we can estimate $p_{fg}$ using its relative frequency by simply counting the occurrences of $\#(f>g)$ and dividing by the total comparisons between $f$ and $g$, $p_{fg} = \frac{\#(f>g)}{\#(f>g) + \#(g>f)}$.

There exist many algorithms to infer the Bradley-Terry strength parameters from this data. Minorization-maximization algorithms provide Maximum Likelihood Estimates (MLE) for the Bradley-Terry strength parameters. However, since the MLE is not guaranteed to be unique for Bradley-Terry models, in this paper we prefer to use the Maximum A Posteriori (MAP) estimate, obtained after assuming uniform priors \cite{BT-MAP,BT-estimation}. Gibbs samplers have been found to work well for MAP parameter estimation of Bradley-Terry models. Individually, the strength parameters $\beta_i$ can be understood to represent the ease of modification of each individual feature. Thus, starting with a set of pairwise comparisons of the ease-of-modification, we can find the MAP estimates of the Bradley-Terry model strength parameters, and consider the additive inverse of these to be the feature costs $\forall f \in \mathcal{F}, Cost(f) = -\beta_f$.

The Bradley-Terry model explicitly yields the probability of a single feature being easier-to-modify than another. This can be extended to represent the probability of an entire set of features together being easier-to-modify than another. Consider two different recourses (sets of features) to modify: $R_1 \subseteq \mathcal{F}$ and $R_2 \subseteq \mathcal{F}$. If $|R_1| = m$ and $|R_2| = n$, and assuming $0 \leq k \leq min(m,n)$ features are common between both recourses, we have $(m-k) \times (n-k)$ pairwise comparisons across the two possible recourses. For each feature pair $f \in R_1, g \in R_2$, we can compute $p_{f>g}$ and use these to compute the corresponding probability that $R_1$ is easier to modify than $R_2$ defined as $\rho_{R_1 > R_2}$ using the following formula (assuming $k = 0$):

\vspace{-0.1in}
\begin{align}
    \rho_{R_1 > R_2} &= \frac{\sum\limits_{f \in R_1, g \in R_2} p_{f>g}}{ \sum\limits_{f \in R_1, g \in R_2} p_{f>g} + \sum\limits_{f \in R_1, g \in R_2} p_{g>f}} \nonumber \\
    &= \frac{\sum\limits_{f \in R_1, g \in R_2} p_{f>g}}{ m \times n } 
\end{align}


Intuitively, $\rho_{R_1 > R_2}$ represents the overall probability that, with $f \in R_1$ and $g \in R_2$, more feature pairs $(f, g)$ are of the form $f>g$ than $g>f$. In practice, this can be easily computed using a Monte Carlo simulation. The case when $k \neq 0$ features are common between $R_1 and R_2$, can be worked out identically, and yields a very similar expression.

\textbf{Importance of Numeric Costs:} We show here that, in order to select between multiple potential recourses, it is insufficient to have access only to pairwise comparisons without explicit numeric costs on every feature. The feature-wise Bradley-Terry strength parameters $\beta_f$ can, however, be used to compare the \textit{overall} ease of modification of two different potential recourses. Thus, this provides us with a way to guide our recourse search algorithms to find the best possible recourses for end users. 
Consider a given data instance $\mathcal{I}$ with corresponding recourse $R$. We define recourse $R$ to be \textit{ideal} if no other recourse $R'$ exists for the same data instance $\mathcal{I}$ such that $\rho_{R>R'} < \rho_{R'>R}$. Intuitively, $R$ thus represents a recourse of minimal cost. Finally, we define a recourse $R$ to be disambiguated if it can be asserted with certainty whether $R$ is \textit{ideal} or \textit{non-ideal}.

\begin{theorem}
Given an order over feature costs, without their exact values, not all possible recourses can be disambiguated.
\end{theorem}
\vspace{-0.2in}
\begin{proof}[Proof]

To show that not all possible recourses can be disambiguated is the same as showing that there exists at least one recourse $R$ that cannot be disambiguated - that is $R$ cannot be shown necessarily to be \textit{ideal} or \textit{non-ideal}. Thus we wish to construct a recourse that is not possible to disambiguate. We can demonstrate that a recourse cannot be disambibuated if there exist two sets of costs $C_1$ and $C_2$ over the same set of features $\mathcal{F}$ such that the features always have the same order, when sorted by cost, and $R$ is determined to be \textit{ideal} using one $C_1$, but \textit{non-ideal} using $C_2$. In this case, since both sets of costs would follow the same order, but differ only in numeric value, it would become necessary to know the numeric value of the costs in order to disambiguate the recourses.

We construct such a scenario using the example recourses in Figure 1. Consider $\mathcal{F} = \{$amt, add, inc, age$\}$, with one set of costs being $C_1 = {-ln(10),-ln(3),-ln(2),-ln(1)}$ for each feature, respectively, and another set being $C_2 = {-ln(10),-ln(9),-ln(8),-ln(1)}$. Even though $C_1 \neq C_2$, the order of the relative order of the feature costs is the same in both cases. We consider Recourse 1 from the figure to be $R$, and Recourse 2 to be $R'$. Thus each recourse modifies two features, $R_1 = \{$amt, age$\}$ and $R_2 = \{$add, inc$\}$. Using feature costs from $C_1$, we can now compute $p_{\text{amt}> \text{add}} = \frac{e^{ln(10)}}{e^{ln(10)} + e^{ln(3)}} = \frac{10}{10+3}$. Similarly, we get $p_{\text{amt}> \text{inc}} = \frac{10}{10+2}$, $p_{\text{inc}> \text{add}} = \frac{1}{3}$, and $p_{\text{age}> \text{inc}} = \frac{1}{1+2}$. These quantities can be used to calculate $\rho_{R > R'} = 0.55$, and similarly $\rho_{R' > R} = 0.45$, that is $\rho_{R > R'} > \rho_{R' > R} \textbf{(1)}$. However, using feature costs from $C_2$, we get $\rho_{R > R'} = 0.32$, and $\rho_{R' > R} = 0.68$, which indicates $\rho_{R > R'} < \rho_{R' > R} \textbf{(2)}$, showing that $R$ is not \textit{ideal}. Equation $\textbf{(1)}$ indicates, however, that provided $R$ and $R'$ are the only possible recourses, $R$ is \textit{ideal}. Since $\textbf{(1)}$ and $\textbf{(2)}$ contradict each other, despite the order between the feature costs being ascending in both $C_1$ and $C_2$, we conclude that order alone is not enough to disambiguate recourses. However, if we know the exact numeric values of the features, we would have been able to disambiguate them in this case.
\end{proof}
\vspace{-0.15in}

\textbf{Comparing Recourses instead of Features:} We have now shown the theoretical need for numeric costs associated with individual features, so that recourse generation algorithms can effectively find meaningful recourses for end users. We have also shown how existing techniques, leveraging MAP estimation algorithms for the Bradley-Terry model, can retrieve these numeric costs when provided with survey data consisting of pairwise comparisons over all possible pairs of features. We now briefly postulate an extension to our proposed solution so far. An underlying assumption we have made is that users will be able to provide meaningful comparisons between all sets of features. This can be extremely hard in real world scenarios. For instance, it can be hard to decide which of two immutable features such as race and national-origin is less easy to modify. Since we are postulating that the Bradley-Terry model applies over features, we believe the latent strength parameters $\beta_f$ exist for every feature $f \in \mathcal{F}$, but that users might find it difficult to actually provide the comparison data for certain features. To make our survey mechanism robust to potentially inane choices that can arise when comparing single features, we propose the comparison of two recourses rather than two features. Thus, instead of asking humans to provide ease-of-modification comparisons $f>g$ or $g>f$ of single features $f \in \mathcal{F}$ and $g \in \mathcal{F}$, we instead ask them provide comparisons $R_1>R_2$ or $R_2>R_1$ between entire recourses (subsets of features) $R_1 \subseteq \mathcal{F}$ and $R_2 \subseteq \mathcal{F}$. Instead of observing $p_{f>g}$ values from our surveys, we thus aim to directly observe $\rho_{R_1>R_2}$ values, and use these to infer individual Bradley-Terry strength parameters for each single feature. When running MAP inference algorithms for the strength parameters $\beta_f$, we parse each occurrence $R_1 > R_2$ in our survey data as $f > g \forall f \in R_1, g \in R_2$. We demonstrate experimentally in the following section that this approach can indeed retrieve individual Bradley-Terry parameters without pairwise comparisons.

\vspace{-0.1in}
\section{Simulations}
\vspace{-0.05in}
In this section we describe the simulations we ran to verify if our approach could learn numeric feature costs from only pairwise comparisons. Instead of surveying real human users, we chose to simulate the data generation process algorithmically. We ran 4 different simulations postulating a Bradley-Terry model over different possible feature sets $\mathcal{F}$ of varying sizes. The strength parameters were drawn uniformly between $0$ and $1$ and used to generate the comparison data consisting of statements of the form $f>g$ for arbitrary $f,g \in \mathcal{F}$. Finally we computed MAP estimates of the strength parameters using only the comparison data and uniform priors, and measured (in terms of Mean Squared Error) how well we were able to recover the initial values of the strength parameters. As is visible from the plot in Figure 2, it is possible to efficiently recover the latent strength parameters for each feature. The entire simulation was run in Python 3.7 on an Intel i7(U) CPU, with runtimes visible in Figure 3.


\vspace{0.05 in}
\begin{figure}[ht]
\begin{center}
\centerline{\includegraphics[width=\columnwidth]{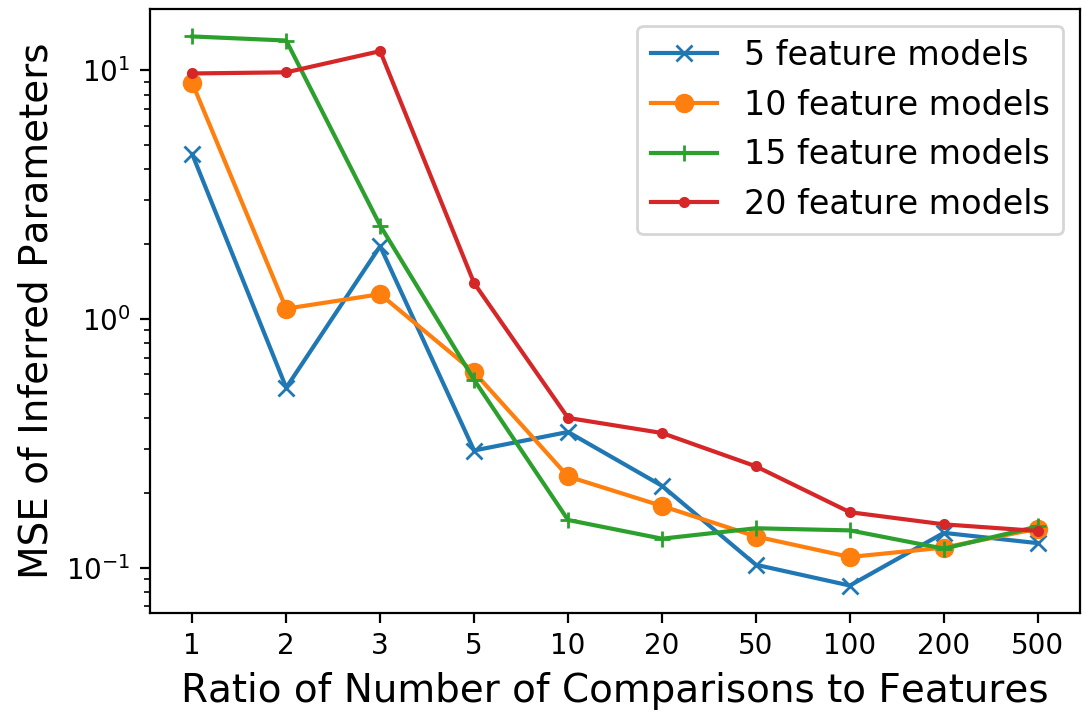}}
\vspace{-0.05 in}
\caption{Mean Squared Error of final estimated Bradley-Terry strength parameters (w.r.t initial) for increasing sizes of survey data (increasing $\frac{Total Number of Comparisons}{|\mathcal{F}|}$). Four different simulations, each with a different size of $|\mathcal{F}|$ ranging from 5 to 20.}
\label{fig2}
\end{center}
\vskip -0.1in
\end{figure}


\vspace{-0.15 in}
\begin{figure}[ht]
\begin{center}
\centerline{\includegraphics[width=\columnwidth]{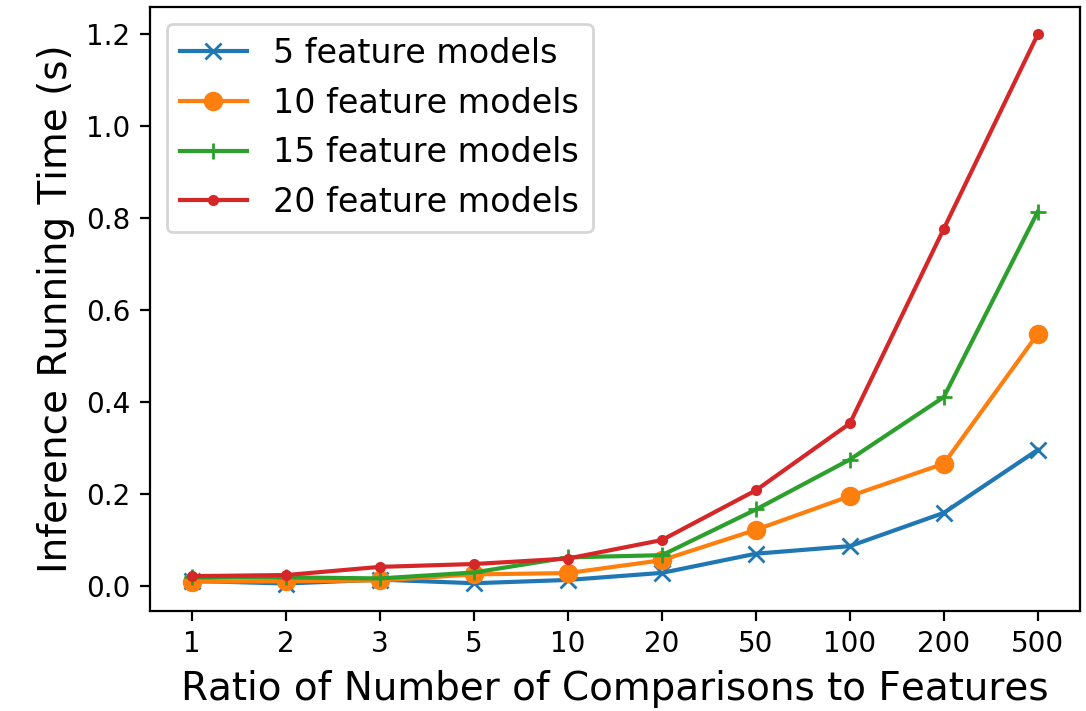}}
\vspace{-0.05 in}
\caption{Running times for the experiment in Figure 2.}
\label{fig3}
\end{center}
\vskip -0.15in
\end{figure}

We ran a different simulation to analyse whether our postulated technique of comparing entire recourses with each other instead of pairs of single features is effective. We considered a feature set of size 20 $|\mathcal{F}| = 20$, and assumed that all the recourses being generated were of a single size, varying from $2$ to $6$. We arbitrarily sampled potential recourses $R \subset \mathcal{F}$ and used equation (2) to determine which recourse was easier to modify overall, constructing a dataset with statements of the form $R_1 > R_2$. We ensured that $R_1 \cap R_2 = \emptyset$. Finally, we parsed each occurrence of $R_1 > R_2$ in our survey data as $f > g \forall f \in R_1, g \in R_2$, and weighed the samples by $\frac{1}{|R_1| \times |R_2|}$. The effectiveness of the parameters retrieved in this simulation can be seen in Figure 4, and corresponding runtimes in Figure 5.

\vspace{0.05 in}
\begin{figure}[ht]
\begin{center}
\centerline{\includegraphics[width=\columnwidth]{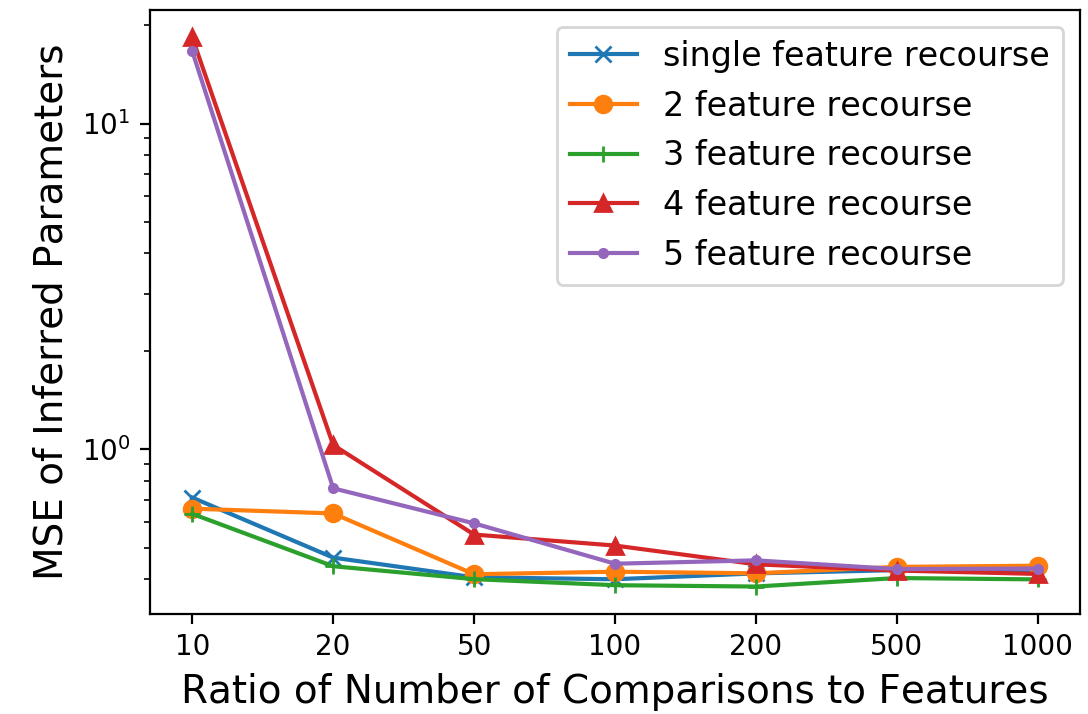}}
\vspace{-0.05 in}
\caption{Mean Squared Error of the final estimated Bradley-Terry strength parameters (w.r.t initial) of a 20 feature model, when survey data consists of comparisons of recourses instead of single features. Recourse sizes simulated range from 1 to 6.}
\label{fig4}
\end{center}
\vskip -0.1in
\end{figure}

\vspace{-0.15 in}
\begin{figure}[ht]
\begin{center}
\centerline{\includegraphics[width=\columnwidth]{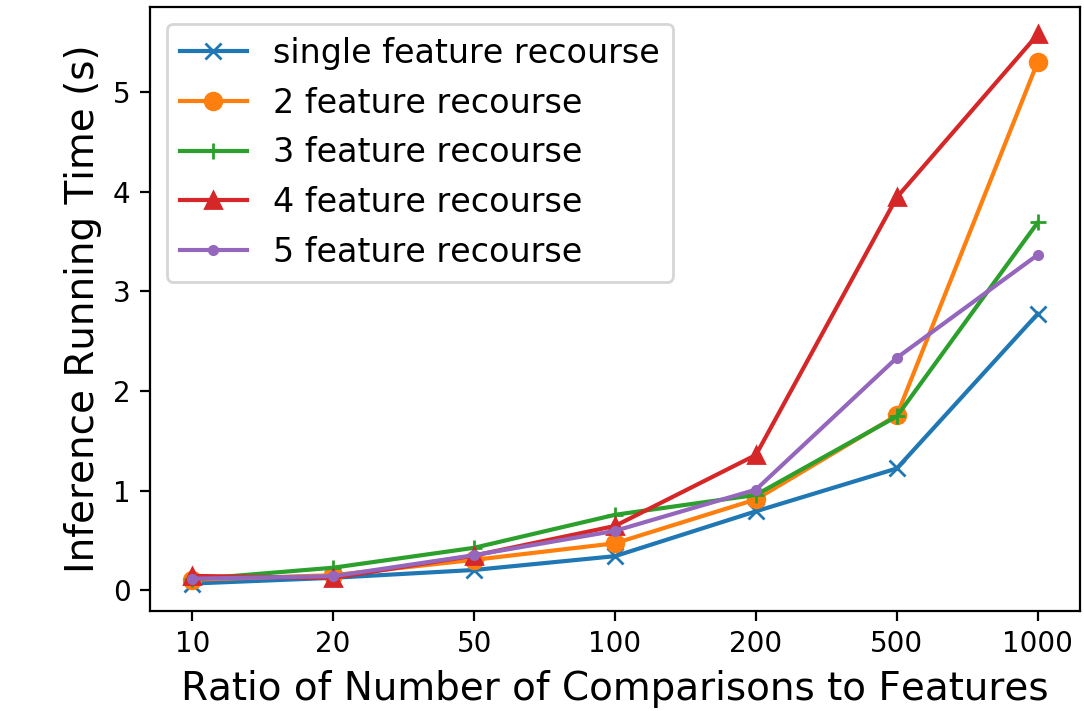}}
\vspace{-0.05 in}
\caption{Running times for the experiment in Figure 4.}
\label{fig5}
\end{center}
\vskip -0.15in
\end{figure}

\vspace{-0.1in}
\section{Conclusion}
\vspace{-0.05in}
In this work we have shown the importance of having numerical costs to measure the ease of modification for each feature, when generating recourse. We have also demonstrated how these costs can be inferred using the Bradley-Terry model, without requiring them to be supplied directly by users. Further, our technique only requires users to indicate which among the finally generated recourses are easier to implement, rather than explicitly compare inane features. While our simulation shows that these comparisons of sets of features are enough to learn costs over individual features, the theoretical proof for this remains to be found as future work.

\clearpage


\bibliography{output}
\bibliographystyle{icml2020}

\appendix

\section{Appendix}

The code used to run the simulations described in this paper can be found at: \url{https://github.com/kaivalyar/RealWorldRecourse}.


\end{document}